\documentclass[10pt,twocolumn,letterpaper]{article}

\usepackage{iccv}
\usepackage{times}
\usepackage{epsfig}
\usepackage{graphicx}
\usepackage{amsmath}
\usepackage{amssymb}

\usepackage{framed}

\usepackage{multirow}
\usepackage{arydshln}
\usepackage{gensymb}
\usepackage{graphicx}
\usepackage{booktabs}

\usepackage{fdsymbol}

\usepackage[breaklinks=true,bookmarks=false]{hyperref}

\iccvfinalcopy 

\def\iccvPaperID{****} 

\ificcvfinal\fi

\begin{document}

\title{Turn That Frown Upside Down: \\
FaceID Customization via Cross-Training Data}

\author{
Shuhe Wang$^{\spadesuit}$, 
Xiaoya Li$^{\blacklozenge}$,
Xiaofei Sun$^{\clubsuit}$ \\ 
Guoyin Wang$^{\blacktriangle}$, 
Tianwei Zhang$^{\blacktriangledown}$,
Jiwei Li$^{\clubsuit}$,
Eduard Hovy$^{\spadesuit}$
}

\maketitle
\ificcvfinal\thispagestyle{empty}\fi

\begin{abstract}
\let\thefootnote\relax\footnotetext{Email: shuhewang@student.unimelb.edu.au}
\let\thefootnote\relax\footnotetext{$^{\spadesuit}$The University of Melbourne, $^{\clubsuit}$Zhejiang University, $^{\blacklozenge}$University of Washington, $^\blacktriangle$01.AI, $^\blacktriangledown$Nanyang Technological University}

Existing face identity (FaceID) customization methods perform well but are limited to generating identical faces as the input, while in real-world applications, users often desire images of the same person but with variations, such as different expressions (e.g., smiling, angry) or angles (e.g., side profile). This limitation arises from the lack of datasets with controlled input-output facial variations, restricting models' ability to learn effective modifications.

%

To address this issue, we propose CrossFaceID, the first large-scale, high-quality, and publicly available dataset specifically designed to improve the facial modification capabilities of FaceID customization models. Specifically, CrossFaceID consists of 40,000 text-image pairs from approximately 2,000 persons, with each person represented by around 20 images showcasing diverse facial attributes such as poses, expressions, angles, and adornments. During the training stage, a specific face of a person is used as input, and the FaceID customization model is forced to generate another image of the same person but with altered facial features. This allows the FaceID customization model to acquire the ability to personalize and modify known facial features during the inference stage. Experiments show that models fine-tuned on the CrossFaceID dataset retain its performance in preserving FaceID fidelity while significantly improving its face customization capabilities.

To facilitate further advancements in the FaceID customization field, our code, constructed datasets, and trained models are fully available to the public.\footnote{Our codes, models, and datasets are available at: \url{https://github.com/ShuheSH/CrossFaceID}.}

\end{abstract}

\section{Introduction}

\begin{figure*}[htb]
\centering
    \includegraphics[scale=0.45]{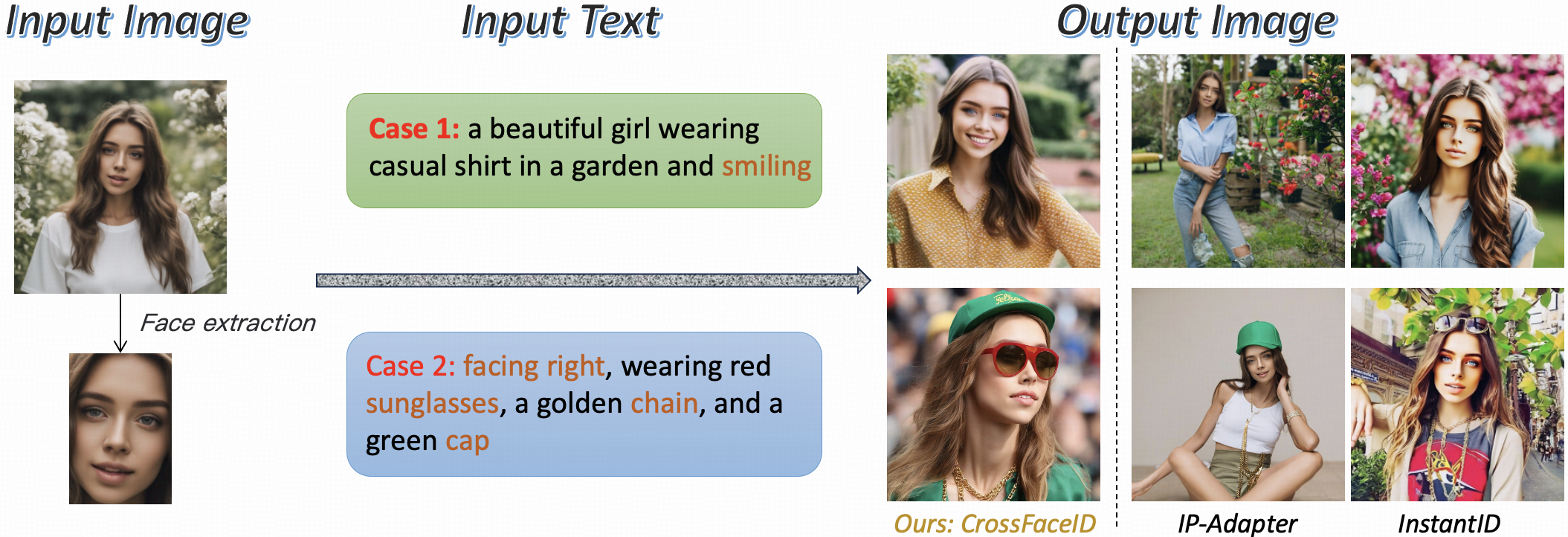}
    \label{fig:intro_example}
    \caption{An example illustrating the limitations of existing FaceID customization models in generating images of the same individual with variations. Here, we use the same image of a girl as the input, paired with two different text prompts. In \textbf{Case 1}, the desired output is an image of the person with smiling, and in \textbf{Case 2}, the goal is an image of the person facing right and wearing sunglasses. The results clearly show that the two leading FaceID customization models, IP-Adapter \cite{ye2023ip} and InstantID \cite{wang2024instantid}, perform well in preserving the exact same face as the input image but fail to customize the input face as specified—such as adding a smile for Case 1 or generating a right-facing person with sunglasses for Case 2. In contrast, the model trained on our proposed CrossFaceID dataset effectively addresses these shortcomings, successfully generating a smiling face for Case 1 and a right-facing person wearing sunglasses for Case 2.}
\end{figure*}

Face identity (FaceID) customization is an important image generation task \cite{nichol2021glide,ramesh2022hierarchical,saharia2022photorealistic,rombach2022high,gal2022image,kumari2023multi,ruiz2023dreambooth,xu2024imagereward}, allowing users to achieve personalized facial customization using a pre-trained text-to-image diffusion model. Although existing methods demonstrate effectiveness, they exhibit a significant limitation: they can only generate images with the exactly same face as the input, while in real-world applications, users often desire images of the same individual but with variations, such as different facial expressions (e.g., smiling) or angles (e.g., side profile), shown in Figure \ref{fig:intro_example}.

%

This issue is primarily due to the lack of such a FaceID customization dataset where input and output faces exhibit controlled variations. 
In current datasets for training face customization models, the input and output faces in the dataset are often identical \cite{kim2022diffface,chen2023photoverse,zhao2023diffswap,peng2024portraitbooth}. This setup restricts the model's ability to learn effective facial feature modifications during training, instead reinforcing its focus on maintaining face consistency. Consequently, the model struggles during inference when it is required to modify facial features while preserving FaceID consistency due to the lack of relevant training experience. 


To address the above issue, in this paper, we propose CrossFaceID, the first large-scale, high-quality, and publicly available dataset specifically designed to improve the facial modification capabilities of FaceID customization models. Specifically, to obtain multiple public images of the same person, CrossFaceID gathered 40,000 images from approximately 2,000 celebrities, with each celebrity represented by around 20 images showcasing diverse facial attributes such as poses, expressions, angles, and adornments. To annotate these images, we use GPT-4 to generate detailed descriptions for these 40,000 images, particularly focusing on the facial features of the individuals, resulting in a set of 40,000 one-to-one text-image pairs.

In this way, during the training stage,  we start with a pre-trained FaceID customization model to ensure a solid foundation for preserving FaceID fidelity. Then, we employ a cross-training method, where a specific face of a person is used as input, and the pre-trained model is forced to generate another image of the same person but with altered facial features. This allows the pre-trained FaceID customization model to acquire the ability to personalize and modify known facial features during the fine-tuning process.

Trained on our CrossFaceID dataset, we achieve comparable performance in preserving FaceID fidelity when compared to the widely used FaceID customization frameworks, InstantID \cite{wang2024instantid} and IP-Adapter \cite{ye2023ip}, while also significantly enhancing their ability to customize FaceID. The contribution of this work can be summarized as follows: 

\begin{itemize}
	\item[1] We collect CrossFaceID, a high-quality, publicly available dataset designed to improve the facial modification capabilities of FaceID customization models.
	\item[2] The model fine-tuned on the CrossFaceID dataset retains its performance in preserving FaceID fidelity while significantly improving its face customization capabilities, showing the effectiveness of the dataset.
	\item[3] Our code, models, and datasets are fully available to the public, supporting further advancements in the FaceID customization field.
\end{itemize}

The rest of this paper is organized as follows: 
\begin{itemize}
	\item Section \ref{sec:related_work} reviews studies related to text-to-image models and FaceID customization.
	\item Section \ref{sec:dataset_construction_crossfaceid} details on the construction of CrossFaceID.
	\item Section \ref{sec:crossfaceid_based_faceid_customization} focuses on the training and inference processes utilizing our constructed CrossFaceID dataset.
	\item Section \ref{sec:experiments} presents experimental results evaluating the effectiveness of CrossFaceID.
	\item Section \ref{sec:conclusion} concludes this work.
\end{itemize}

\section{Related Work}
\label{sec:related_work}

\subsection{Text-to-image Models}
Text-to-image refers to the process of generating images from textual descriptions using pre-trained image generation models \cite{ramesh2021zero,ding2021cogview,ding2022cogview2,ramesh2022hierarchical,rombach2022high,saharia2022photorealistic,huang2023composer}.
These models are trained to understand the relationship between textual input and visual content, enabling them to create images that match the given description.
Thanks to the success of the transformer model \cite{vaswani2017attention}, most early text-to-image approaches can be broken down into two stages: (1) using an image encoder, such as DARN~\cite{gregor2014deep}, PixelCNN~\cite{van2016conditional}, PixelVAE~\cite{gulrajani2016pixelvae}, or VQ-VAE \cite{van2017neural}, to convert an image into several tokens; and (2) training the model to predict these image tokens based on the provided text input within the transformer framework \cite{vaswani2017attention}. 
Recently, diffusion models \cite{song2020denoising,song2020score,nichol2021glide,dhariwal2021diffusion,ramesh2022hierarchical,saharia2022photorealistic,rombach2022high,balaji2022ediff,huang2023composer} have emerged as the new state-of-the-art approach for image generation, offering innovative solutions for the text-to-image task.
In this approach, the text prompt is first encoded into embeddings using a pre-trained language model such as T5 \cite{raffel2020exploring} or CLIP \cite{radford2021learning}, and then these encoded embeddings are used to guide the diffusion process, resulting in the generation of high-quality images.
For example, GLIDE \cite{nichol2021glide} employs a cascaded diffusion architecture with CLIP \cite{radford2021learning} as the text encoder to condition on natural language descriptions, facilitating both image generation and editing. Imagen \cite{saharia2022photorealistic} adopts T5 \cite{raffel2020exploring}, a generic large language model pre-trained on text-only corpora, as the text encoder of diffusion models, to further enhance the text understanding.

\subsection{ID Customization}

FaceID customization for text-to-image models refers to the process of personalizing the text-to-image generation model by tailoring it to better recognize, and generate facial features and attributes specific to individual users \cite{valevski2023face0,ye2023ip,yuan2023inserting,chen2024dreamidentity,wang2024instantid,xiao2024fastcomposer,li2024photomaker,peng2024portraitbooth}.
Most of these works are optimization-free methods, which directly encode FaceID information into the generation process. 
For instance, Face0 \cite{valevski2023face0} substitutes the last three text tokens with the projected face embedding in the CLIP \cite{radford2021learning} space, using the resulting combined embedding to guide the diffusion process. In a similar vein, PhotoMaker \cite{li2024photomaker} adopts a related approach but enhances its ability to extract FaceID embeddings by fine-tuning specific Transformer \cite{vaswani2017attention} layers in the image encoder and merging the class and image embeddings. Additionally, IP-Adapter \cite{ye2023ip} and InstantID \cite{wang2024instantid} leverage FaceID embeddings from a face recognition model rather than CLIP image embeddings, ensuring consistent ID representation.
However, these methods primarily concentrate on improving FaceID fidelity while overlooking customization. As illustrated in Figure \ref{fig:intro_example}, due to the complex architecture required to preserve FaceID, it is challenging for them to customize the generated face simply by modifying the input prompt. For example, generating an image of a person smiling when the input face does not show a smile becomes difficult.
In this paper, we address this issue by altering only the composition of the training data, without changing the model structure. Extensive experiments demonstrate that our approach can effectively modify or preserve the input FaceID based on the input text prompt.

%
%
%
%
%
%
%
%
%

\begin{figure*}[htb]
\centering
    \includegraphics[scale=0.96]{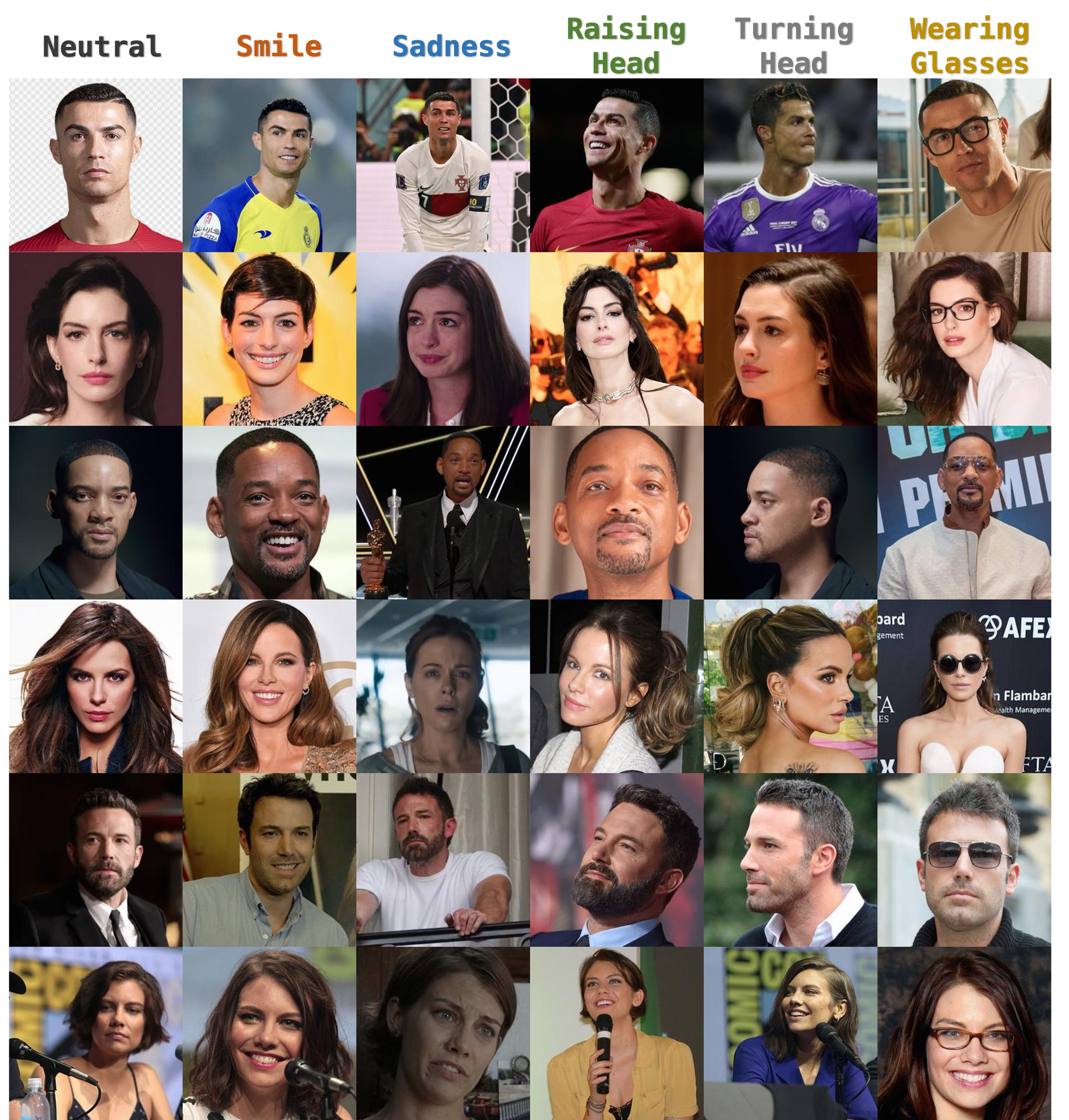}
    \label{fig:example_celebrity}
    \caption{Images sampled from our constructed CrossFaceID dataset are presented. From top to bottom, the display includes six persons, each occupying one row with six images showing different expressions and angles: \textbf{(1) No Expression}: a straight-faced person with no expression, \textbf{(2) Smile}: a person with a smiling face, \textbf{(3) Sad}: a person with a sad expression, \textbf{(4) Rise}: a person with his or her face rising up, \textbf{(5) Side}: a person with his or her face turned to one side, and \textbf{(6) Wearing Glasses}: a person wearing glasses.}
\end{figure*}

\section{Dataset Construction: CrossFaceID}
\label{sec:dataset_construction_crossfaceid}

In this section, we provide details on the dataset construction of CrossFaceID, which aims to address the issue in existing FaceID customization datasets, where input and output faces do not exhibit controlled variations.
 The construction process starts with images of the same individual. Next, we gather multiple images that showcase the person under varying facial attributes, such as different expressions, angles, or adornments. Using GPT-4 \cite{achiam2023gpt}, we then generate detailed textual descriptions for each image, particularly focusing on the facial features of the individual, resulting in a dataset of one-to-one text-image pairs.


%
%


\subsection{Image Collection}
To gather multiple public images of the same individual, we focus on celebrities, as they have numerous publicly available images on the Internet showcasing variations in facial features. As a result, we crawled approximately 60K images from 1626 celebritys.

\subsection{Image Filtering}

The collected images from the Internet may include noise, such as blurred faces or low resolution, which could significantly impact the quality of the subsequent FaceID customization models. To address this issue, we manually established rules to carefully select high-quality images suitable for FaceID customization training:

\textbf{(1) The image must include faces, with a maximum of three faces allowed.}
This condition is based on two key reasons: firstly, when an image contains multiple faces, the model may have difficulty identifying which face to focus on, potentially mixing up facial features between individuals. Secondly, during the inference stage, the trained FaceID customization model is typically designed to work with a single primary face. Limiting the number of faces ensures consistency with the inference process.

\textbf{(2) The image resolution must be at least 512x512 or higher.}
High-resolution images generally contain finer facial details, such as subtle expressions, skin texture, and small features like wrinkles or dimples. As a result, they offer richer visual information for the FaceID customization model to analyze, leading to improved feature extraction and more effective learning.

\textbf{(3) The face should be at least 4\% of the image.}
This is because larger facial regions provide more pixels dedicated to facial features, enabling the FaceID customization model to better capture details such as expressions, and facial textures, which are critical for FaceID customization. As for the number ``4\%", it was determined through iterative refinements and validated via human and model evaluations.

The detailed statistics for the remaining 
dataset
are shown in Table \ref{tab:statistics_crossfaceid}.
After cleaning, we ensure that the collected images include faces that are clear and of adequate size. Examples of cleaned images are shown in Figure \ref{fig:example_celebrity}. 

\begin{table}[t]
	\centering

	\begin{tabular}{lc}
	\toprule
	
	Statists & Number \\\midrule
	Total Celebrities & 1626 \\
	Total Images & 40596 \\
	Average Images Per Celebrity & 24.9 \\
	Medium Images Per Celebrity & 7 \\
	Maximum Images Per Celebrity & 111 \\
	Minimum Images Per Celebrity & 2 \\\bottomrule
	\end{tabular}

	\caption{The image statistics for the cleaned CrossFaceID dataset.}
	\label{tab:statistics_crossfaceid}
\end{table}

\subsection{Image Annotations with GPT-4o}

Since the crawled images lack captions, which are essential for training FaceID customization models, we leverage GPT-4o \cite{hurst2024gpt} to annotate these images. This annotation process generate detailed textual descriptions with a specific emphasis on the individual's facial features.

Given a person image, we prompt GPT-4o \cite{hurst2024gpt} to generate detailed textual descriptions. The designed prompt can be decomposed into the following three aspects:



(1) Task overview, 

``\textit{Your goal is to using less than 4 fluent sentences to generate short, descriptive captions for this image.}"

\noindent which emphasizes that the task is an annotation task, and the generated captions should be concise and not overly lengthy to avoid negatively impacting the latter training process.

(2) Instructions on face description,

``\textit{Note that your description should be as detailed as possible to describe the features of the face in the image.}"

\noindent which instructs GPT-4o to focus on providing detailed descriptions of the facial features in the image within the constraints of the output length. It should avoid elaborating extensively on non-essential background information, such as character actions, attire, or surroundings, as these are not central to our following FaceID customization task.

(3) Precautions,

``\textit{Please note that your answer should not include any tabulation format.}"

\noindent which prevents GPT-4 from generating structured descriptions. This is due to our observations in experiments where GPT-4 tended to create structured outputs. For instance, when given the following image:

\begin{figure}[htb]
\centering
    \includegraphics[scale=0.35]{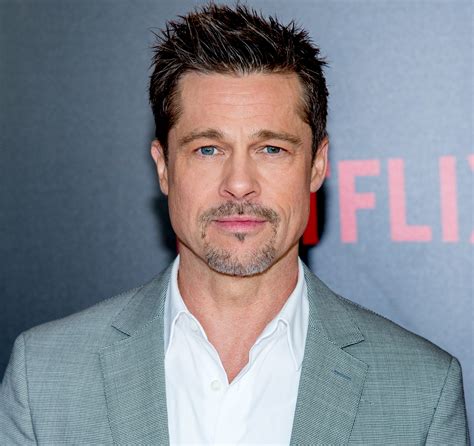}
\end{figure}

\noindent the image description that GPT-4o outputs is as follows:
\begin{framed}
\textbf{1. Hair}: Short and neatly styled brown hair with a slightly spiked texture on the top, adding volume.

\textbf{2. Eyes:} Bright blue eyes framed by well-groomed eyebrows, giving an intense and focused look.   

\textbf{3. Nose:} Straight and proportionate nose, complementing the facial symmetry.

\textbf{4. Mouth:} Neutral expression with closed lips, subtly emphasizing the jawline.

\textbf{5. Facial Hair:} Trimmed goatee and short mustache, enhancing the rugged yet polished appearance.

\textbf{6. Expression:} Confident and composed, with a slight hint of seriousness, fitting a professional or formal setting.

\end{framed}
\noindent which provides a quick overview of the image's content, but it is not ideal for training FaceID customization models. During inference, users are more likely to provide free-form descriptions rather than structured text. Training with structured descriptions could make it difficult for the model to effectively interpret unstructured, natural language inputs.

Then, we use the above designed prompt to interact with GPT-4o \cite{hurst2024gpt} for annotating the crawled images. The parameters for this process are configured as follows:

\textit{model = gpt-4o}, \textit{temperature = }0, \textit{top\_p = }1, \textit{frequency\_penalty = }0, \textit{presence\_penalty = }0, \textit{max\_token = }4096

\begin{figure}[htb]
\centering
    \includegraphics[scale=0.24]{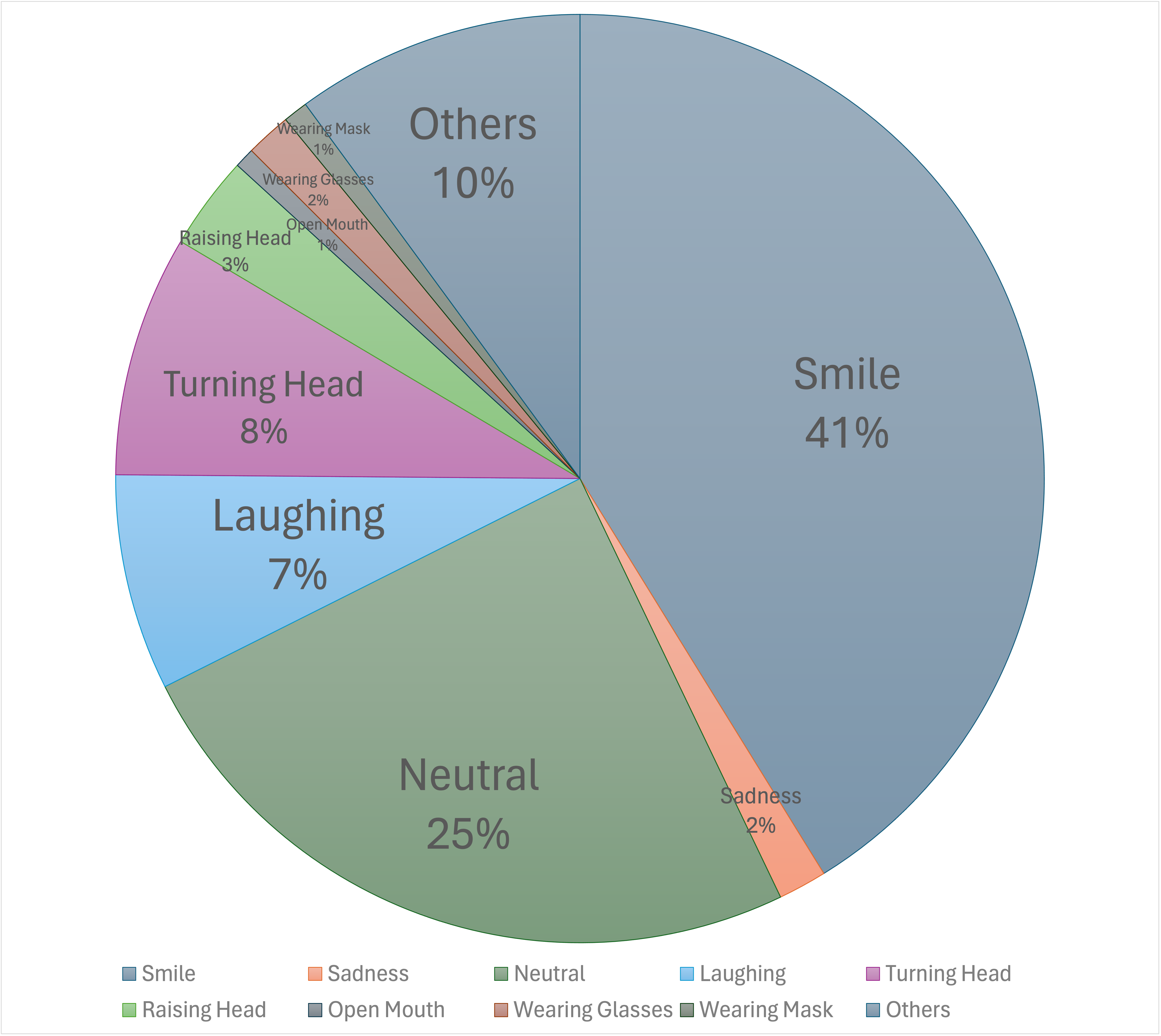}
    \caption{The distribution of various facial features (e.g., expressions and angles) within our CrossFaceID dataset.}
    \label{fig:statistics}
\end{figure}

Figure \ref{fig:statistics} shows the distribution of various facial features (e.g., smile, neutral and wearing glasses) within our CrossFaceID dataset. Next, we will describe how to utilize these data to fine-tune FaceID customization models and perform inference using the trained model.

\section{CrossFaceID Based FaceID Customization}
\label{sec:crossfaceid_based_faceid_customization}

In this section, we first provide an overview of the fundamental concepts for training face ID customization models in Section \ref{sec:preliminaries}. Following that, we delve into the training and inference processes utilizing our constructed CrossFaceID dataset in Section \ref{sec:sec_crossfaceid_based_faceid_customization}.

\subsection{Preliminaries}
\label{sec:preliminaries}


\subsubsection{Diffusion Models}
\label{sec:diffusion_models}

Diffusion models \cite{ho2020denoising,rombach2022high,podell2023sdxl} are a class of generative models that have gained prominence for their ability to generate high-quality, realistic data, such as images, by simulating a gradual process of transforming random noise into structured data.
Specifically, during each training process, noise $\epsilon$ is sampled and added to the input image $x_{0}$ based on a noise schedule (i.e., Gaussian noise). This process yields a noisy sample $x_{t}$ at timestep $t$:

\begin{equation}
x_{t} = \alpha_{t}x + \sigma_{t}\epsilon, \epsilon \sim \mathcal{N}(0, \textbf{I})
\end{equation}
where $\alpha_{t}$ and $\sigma_{t}$ are the coefficients of the adding noise process, essentially representing the noise schedule.
Then the diffusion model $\epsilon_{\theta}$ is forced to predict the normally-distributed noise $\epsilon$ with current added noisy $x_{t}$, time step $t$, and condition information $C$, where commonly $C$ represents the embedded text prompt. For optimization process:

\begin{equation}
\mathcal{L}(\theta) = \mathbb{E}_{x_{0},C,t,\epsilon\sim\mathcal{N}(0,\textbf{I})}\lVert\epsilon - \epsilon_{\theta}(x_{t}, C, t)\rVert^{2}
\end{equation}
where $t\in[0, T]$ is the sampled diffusion step.

The inference stage begins with a sample of pure noise (Gaussian noise), represented by $x_{T}$, where $T$ is a predefined number of timesteps. This random noise is the initial state, representing a completely unstructured and meaningless input.
Then, for each timestep $t$, the model takes the noisy image $x_{t}$ at step $t$ as the input, and incorporates the text prompt as the condition $C$ to predict the clean image or the noise that should be removed to get closer to the final clean image $x_{0}$. The predicted noise $\epsilon_{\theta}$ is then used to update the noisy image, denoising it step by step:
\begin{equation}
x_{t-1} = \alpha_{t}x_{t} - \sigma_{t}\epsilon_{\theta}(x_{t},C,t)
\end{equation}
where $\alpha_{t}$ and $\sigma_{t}$ are two coefficients controlling the denoising process. Finally, over several timesteps $T$, the noise is gradually removed, resulting in a high-quality image.

\subsubsection{IP-Adapter}

IP-Adapter \cite{ye2023ip} is a method for enabling image prompt capabilities alongside text prompts, without altering the original text-to-image models. It utilizes a distinct decoupled cross-attention mechanism, embedding image features through several extra cross-attention layers, while keeping the other model parameters intact.
Specifically, in original diffusion models, the text features from the CLIP \cite{radford2021learning} or T5 \cite{raffel2020exploring} text encoder are incorporated into the model by inputting them into the cross-attention layers. Given the latent image features $Z$ and the text features $C_{text}$, the output of cross-attention $Z^{'}$ can be expressed as:
\begin{equation}
\begin{aligned}
&Z^{'} = \text{Attention}(Q, K, V) = \text{Softmax}(\frac{QK^{T}}{\sqrt{d}})V, \\
&Q = ZW_{q}, K = C_{text}W_{k}, V = C_{text}W_{v}
\end{aligned}
\end{equation}
where $Q$, $K$, and $V$ represent the query, key, and value matrices in the attention operation, respectively, while $W_{q}$, $W_{k}$, and $W_{v}$ are the weight matrices of the learnable layers.

To further incorporate face ID, IP-Adapter \cite{ye2023ip} adds a new cross-attention layer for each cross-attention layer in the original diffusion model. Similarly, given the face ID features $C_{id}$, the output of ID  cross-attention $Z^{''}$ is:
\begin{equation}
\begin{aligned}
&Z^{''} = \text{Attention}(Q, K^{'}, V^{'}) = \text{Softmax}(\frac{Q(K^{'})^{T}}{\sqrt{d}})V^{'}, \\
&Q^{'} = ZW^{'}_{q}, K^{'} = C_{id}W^{'}_{k}, V^{'} = C_{id}W^{'}_{v}
\end{aligned}
\end{equation}
where $Q^{'}$, $K^{'}$, and $V^{'}$ represent the query, key, and value matrices in the attention operation, respectively, while $W^{'}_{q}$, $W^{'}_{k}$, and $W^{'}_{v}$ are the weight matrices of the learnable layers.

The text cross-attention $Z^{'}$ and the ID cross-attention $Z^{''}$ are added, resulting the final decoupled attention $Z^{final}$:

\begin{equation}
\begin{aligned}
Z^{final} &= Z^{'} + \lambda\cdot Z^{''} \\
 &= \text{Attention}(Q, K, V) + \lambda\cdot\text{Attention}(Q, K^{'}, V^{'}) \\
 &= \text{Softmax}(\frac{QK^{T}}{\sqrt{d}})V + \lambda\cdot\text{Softmax}(\frac{Q(K^{'})^{T}}{\sqrt{d}})V^{'}
\end{aligned}
\end{equation}
where $\lambda$ is a weight factor.

During the training stage, IP-Adapter only optimizes the related linear layers within the decoupled cross-attention while keeping the parameters of the diffusion model fixed:

\begin{equation}
\mathcal{L} _{IP}(\theta) = \mathbb{E}_{x_{0},C_{text},C_{id},t,\epsilon\sim\mathcal{N}(0,\textbf{I})}\lVert\epsilon - \epsilon_{\theta}(x_{t}, C_{text}, C_{id}, t)\rVert^{2}
\end{equation}

For the inference phase, the IP-Adapter progressively denoises the input noise to generate the customized face image, following the same process as the basic diffusion models described in Section \ref{sec:diffusion_models}, so we do not reiterate it here.

\begin{figure*}[htb]
\centering
    \includegraphics[scale=0.73]{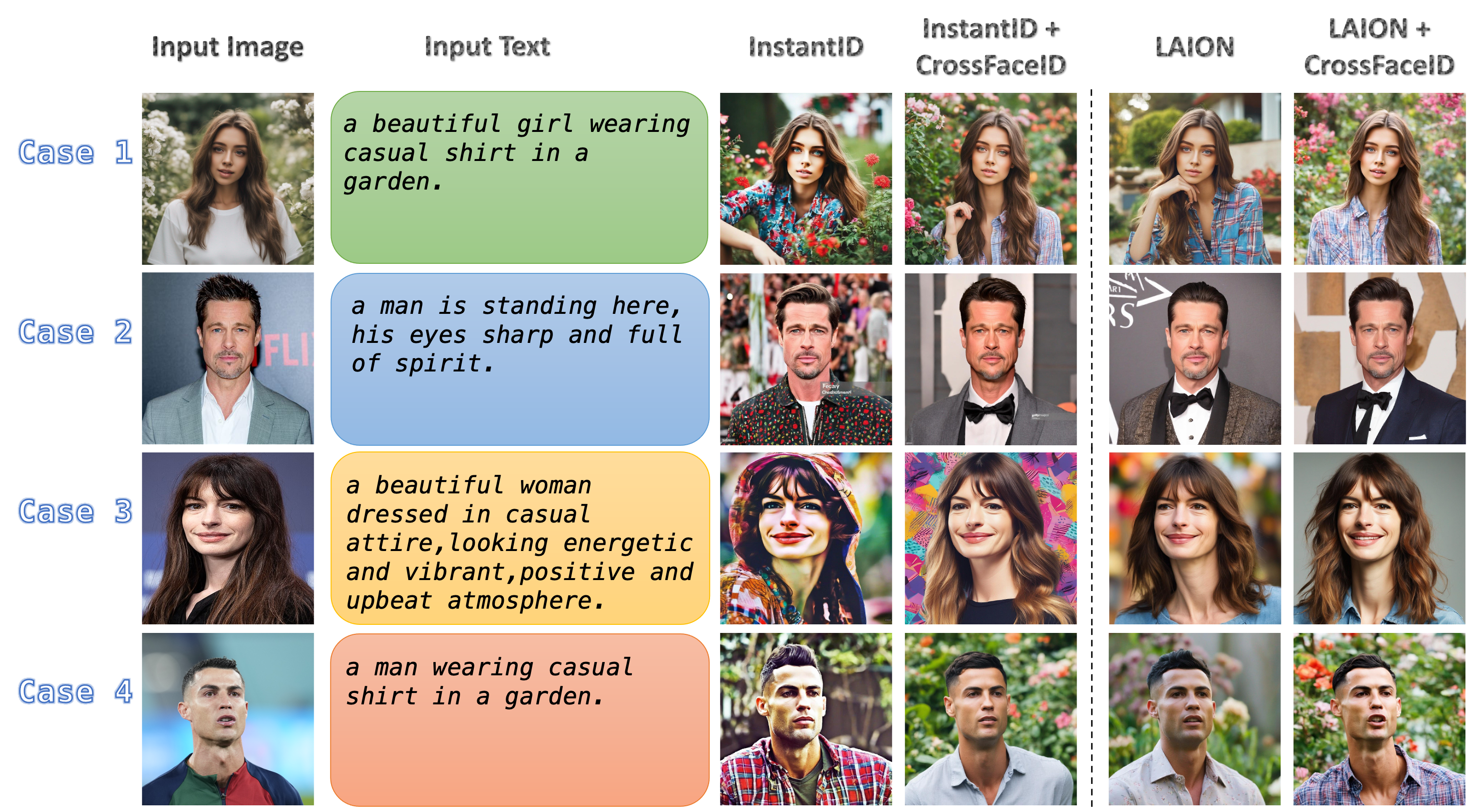}
    \caption{The results demonstrate the performance of FaceID customization models in maintaining FaceID fidelity. For models, ``InstantID" refers to the official InstantID model, while ``InstantID + CrossFaceID" represents the model further fine-tuned on our CrossFaceID dataset. ``LAION" denotes the InstantID model pre-trained on our curated LAION dataset, and ``LAION + CrossFaceID" refers to the model further trained on the CrossFaceID dataset. These results indicate that (1) for both the official InstantID model and the LAION-trained model, the ability to maintain FaceID fidelity remains consistent before and after fine-tuning on our CrossFaceID dataset, and (2) the model trained on our curated LAION dataset achieves comparable performance to the official InstantID model in preserving FaceID fidelity.}
    \label{fig:results_fidelity}
\end{figure*}

\subsubsection{InstantID}

InstantID \cite{wang2024instantid} is an improved version of IP-Adapter \cite{ye2023ip}, designed to generate customized images with different poses or styles based on a face ID image, while maintaining high fidelity. It can be decomposed into three key components: (1) FaceID Embedding, which extracts facial information using advanced visual models; (2) Spatial Facial Information Extraction, which encodes fine-grained features from the face image as supplementary spatial facial data to improve the face ID embedding; and (3) Image Adapter, which utilizes cross-attention similar to IP-Adapter to combine the face ID embedding with the text embedding.

\paragraph{FaceID Embedding.} For a given image, a pre-trained face model is used to detect the face and encode it as the face ID embedding.  This encoded face ID embedding serves as a supplementary input to the text embeddings, helping guide the diffusion model in generating the desired image.

\paragraph{Spatial Facial Information Extraction.}
In contrast to IP-Adapter \cite{ye2023ip}, which combines coarse-grained face information with text prompts, InstantID \cite{wang2024instantid} utilizes five facial key points (two for the eyes, one for the nose, and two for the mouth) as supplementary facial features to refine the encoded face ID embedding.
To achieve this goal, InstantID \cite{wang2024instantid} first employs a pre-trained face model to extract five key points (two for the eyes, one for the nose, and two for the mouth) from the input image as spatial control signals. These signals are then encoded using a ControlNet \cite{zhang2023adding}, which enhances the extracted face ID embedding. 
By integrating spatial facial information, this approach minimizes the influence of spatial constraints and avoids overemphasizing redundant features, such as face shape or mouth closure, thereby preserving the editability of the image.

\paragraph{Image Adapter.}
InstantID \cite{wang2024instantid} adopts the approach of IP-Adapter \cite{ye2023ip} by utilizing decoupled cross-attention to combine text prompts with the enhanced FaceID embedding:

\begin{equation}
\text{Final Attention} = \text{Attention(Text)} + \lambda\cdot\text{Attention(FaceID)}
\end{equation}
where $\lambda$ is a weight factor.

During the training phase, InstantID \cite{wang2024instantid} employs the same approach as IP-Adapter \cite{ye2023ip}, optimizing only the relevant linear layers within the decoupled cross-attention and the ControlNet \cite{zhang2023adding} used for encoding spatial facial information, while leaving the parameters of the pre-trained diffusion model unchanged:

\begin{equation}
\mathcal{L} _{In}(\theta) = \mathbb{E}_{x_{0},C_{text},C_{id},t,\epsilon\sim\mathcal{N}(0,\textbf{I})}\lVert\epsilon - \epsilon_{\theta}(x_{t}, C_{text}, C_{id}, t)\rVert^{2}
\end{equation}
While during the inference phase, it iteratively denoises the input noise to generate the customized face image, which is the same as we detailed in Section \ref{sec:diffusion_models}.

\subsection{FaceID Customization on CrossFaceID }
\label{sec:sec_crossfaceid_based_faceid_customization}

\subsubsection{Training}

During the training phase, we keep the training structure of FaceID customization unchanged as described in Section \ref{sec:crossfaceid_based_faceid_customization}, while modifying the arrangement of the input and output.

Formally, supposed that the collected dataset includes $N$ persons, with each person represented by $n$ triples ($y_{image}, y_{text}, y_{face}$) triples, where $y_{image}$ denotes an image of the person,  $y_{text}$ represents the corresponding text description, and $y_{face}$ refers to the extracted face from $y_{image}$.
Firstly, we use one image of a person, $y_{image}^{i}, 0\leq i \leq n$, as the input image $x_{0}$, and its corresponding text description $y_{text}^{i}$ as the text condition $C_{text}$. However, for the face condition $C_{id}$, we do not use its corresponding face $y_{face}^{i}$. Instead, we select a different random face of the same person, $y_{face}^{j}$, where $j\neq i$. 
Then, the subsequent training step follows the same procedure as standard FaceID customization models detailed in Section \ref{sec:preliminaries}.
Noise $\epsilon$ is sampled and added to the input image $x_{0}$ according to a predefined noise schedule (e.g., Gaussian noise), resulting in a noisy sample $x_{t}$ at timestep $t$.
The diffusion model $\epsilon_{\theta}$ is then trained to predict the normally distributed noise $\epsilon$ using the current noisy image $x_{t}$, the timestep $t$, the text condition $C_{text}$, and the face condition $C_{id}$. For optimization process:

\begin{equation}
\mathcal{L}(\theta) = \mathbb{E}_{x_{0},C_{text},C_{id},t,\epsilon\sim\mathcal{N}(0,\textbf{I})}\lVert\epsilon - \epsilon_{\theta}(x_{t}, C_{text}, C_{id}, t)\rVert^{2}
\end{equation}
where $t\in[0, T]$ is the sampled diffusion step.

For settings, we follow the approach outlined in InstantID \cite{wang2024instantid} focusing on single-person images and utilizing a pre-trained face model, Antelopev2\footnote{\url{https://github.com/deepinsight/insightface}}, to detect and extract face ID embeddings from human images. During training, only the parameters of the Image Adapter and IdentityNet are updated, while the pre-trained text-to-image model remains frozen. Our experiments are conducted using the SDXL-1.0 model on 16 NVIDIA H800 GPUs (80GB) with a batch size of 2 per GPU.

\subsubsection{Inference}

The inference process follows the same approach as the diffusion models outlined in Section \ref{sec:diffusion_models}.
It begins with a sample of Gaussian noise, represented by $x_{T}$, where $T$ is a predefined number of timesteps. This initial state, composed entirely of unstructured noise, serves as the starting point, representing a meaningless input image.
At each timestep $t$, the model takes the noisy image $x_{t}$ as the input and utilizes the text prompt condition $C_{text}$ and the input face condition $C_{id}$ to predict the clean image or the noise that should be removed, progressively refining the image towards the final clean output $x_{0}$. The predicted noise $\epsilon_{\theta}$ is then used to update the noisy image, denoising step by step:
\begin{equation}
x_{t-1} = \alpha_{t}x_{t} - \sigma_{t}\epsilon_{\theta}(x_{t},C_{text}, C_{id},t)
\end{equation}
where $\alpha_{t}$ and $\sigma_{t}$ are two coefficients controlling the denoising process. Over several timesteps $T$, the noise is gradually removed, ultimately producing a customized, clean image.

\begin{figure*}[htb]
\centering
    \includegraphics[scale=0.73]{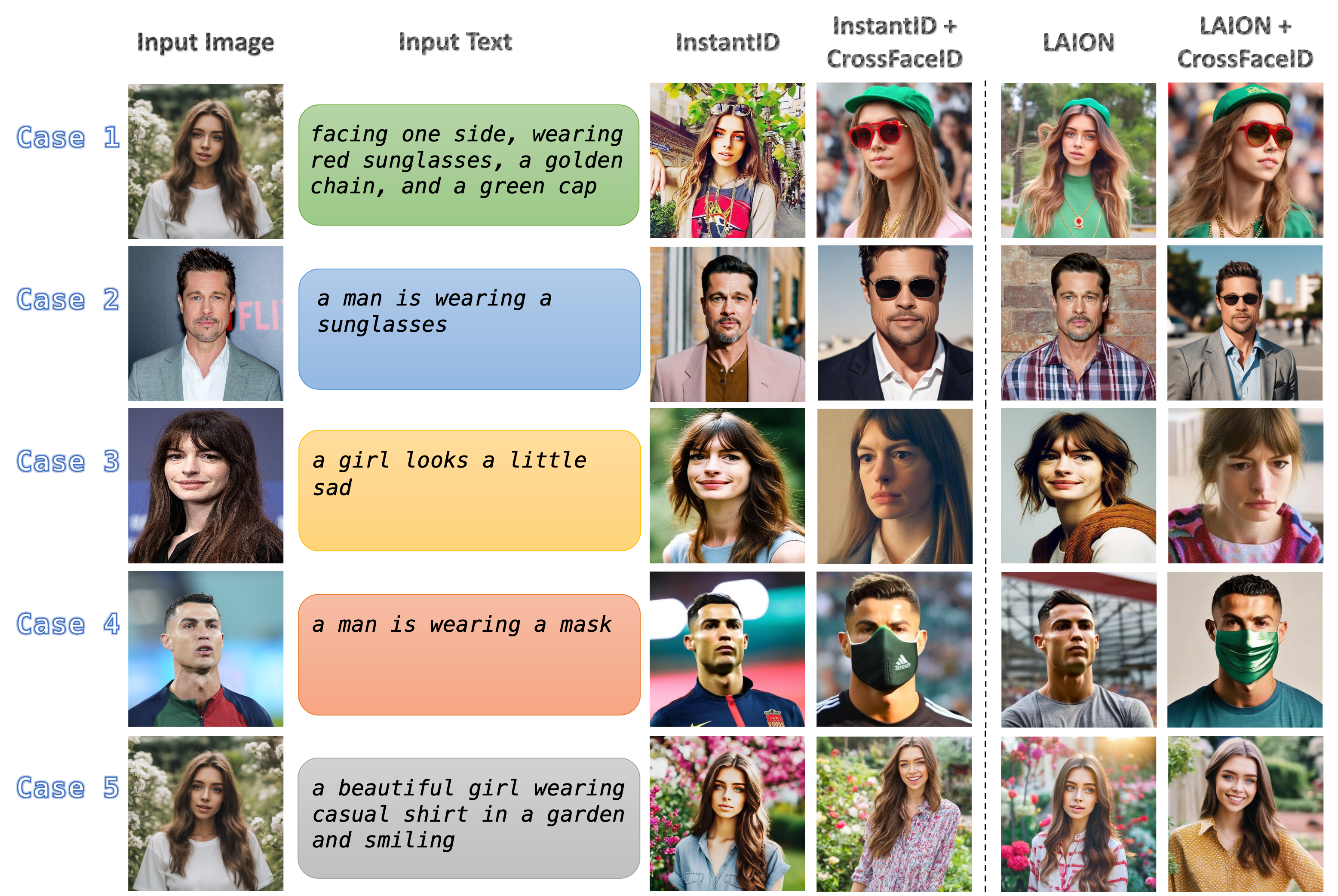}
    \caption{The results of the performance for FaceID customization models in customizing or editing FaceID. Here, "InstantID" represents the official InstantID model, while "InstantID + CrossFaceID" refers to the model fine-tuned on our CrossFaceID dataset. Similarly, "LAION" denotes the InstantID model pre-trained on our curated LAION dataset, and "LAION + CrossFaceID" refers to the model further fine-tuned on the CrossFaceID dataset. From these results, we can clearly observe an improvement in the models' ability to customize FaceID after being fine-tuned on our constructed CrossFaceID dataset.}
    \label{fig:results_customization}
\end{figure*}

\section{Experiments}
\label{sec:experiments}


To assess the effectiveness of CrossFaceID, we train the state-of-the-art FaceID customization model, InstantID \cite{wang2024instantid}, as well as its original version, IP-Adapter \cite{ye2023ip}, using our constructed CrossFaceID dataset. We then perform comparative experiments to evaluate both FaceID fidelity and FaceID customization capabilities. 

\subsection{Main Results}

In this section, we conduct experiments to separately evaluate the FaceID fidelity and FaceID customization capabilities of models trained on our constructed CrossFaceID dataset. For clarity, we refer to the official InstantID model as ``InstantID," and the model further fine-tuned on our CrossFaceID dataset as ``InstantID + CrossFaceID." Similarly, the InstantID model pre-trained on our curated LAION dataset is referred to as ``LAION," while the model further trained on our CrossFaceID dataset is called ``LAION + CrossFaceID."
 Below are results showing their abilities in FaceID fidelity and FaceID customization.

\subsubsection{FaceID Fidelity}

Figure \ref{fig:results_fidelity} illustrates the performance of FaceID customization models in maintaining FaceID fidelity. From these results, we can conclude that (1) for both the official InstantID model and the LAION-trained model, the ability to maintain FaceID fidelity remains consistent before and after fine-tuning on our CrossFaceID dataset. Such as the case 1, all models, including the two fine-tuned on the CrossFaceID dataset generate the exact same girl as the input image and wearing shirt in a garden as the input text ``a beautiful girl wearing casual shirt in a garden".
This demonstrates that our constructed CrossFaceID dataset does not compromise the FaceID fidelity performance of these FaceID customization models. 
(2) 
The model trained on our curated LAION dataset demonstrates performance comparable to the official InstantID model in maintaining FaceID fidelity. For instance, in case 2, both the official InstantID model and the LAION-trained model successfully generate the desired images based on the input.
This ensures the fairness of our experiments when further fine-tuning CrossFaceID on models with comparable baseline performance.

\subsubsection{FaceID Customization}

Figure \ref{fig:results_customization} demonstrates the performance for FaceID customization models in customizing or editing FaceID. From these results, we can clearly observe an improvement in the models' ability to customize FaceID after being fine-tuned on our constructed CrossFaceID dataset. For example, in case 3, although the input image features a smiling girl, both CrossFaceID models, ``InstantID + CrossFaceID" and ``LAION + CrossFaceID," successfully generate images of the girl without a smile, appearing slightly sad as specified by the input text, ``a girl looks a little sad." Moreover, the two CrossFaceID models effectively customize the input person by generating a man wearing sunglasses for case 2 (``a man is wearing sunglasses") and a man wearing a mask for case 4 (``a man is wearing a mask").

\begin{table*}[t]
	\centering

	\begin{tabular}{lcccccc}
	\toprule
	\multirow{2}{*}{\textbf{Model}} & \multicolumn{2}{c}{\textbf{CrossFaceID-test}} & \multicolumn{3}{c}{\textbf{Unsplash-50}} \\
	 & CLIP-T$\uparrow$ & CLIP-I$\uparrow$ & Face Sim$\uparrow$ & CLIP-T$\uparrow$ & CLIP-I$\uparrow$ \\\midrule
	IP-Adapter \cite{ye2023ip} & 0.19 & 0.62 & 0.57 & 0.24 & 0.61 \\
	InstantID \cite{wang2024instantid} & 0.24 & 0.69 & 0.61 & 0.27 & 0.68 \\
	LAION & 0.25 & 0.71 & 0.62 & 0.28 & 0.70 \\\midrule
	\multicolumn{7}{c}{\textit{Ours (Fine-tuned on CrossFaceID)}} \\
	IP-Adapter + CrossFaceID & 0.25 (+0.04) & 0.67 (+0.05) & 0.59 (+0.02) & 0.27 (+0.03) & 0.67 (+0.06) \\
	InstantID + CrossFaceID  & 0.30 (+0.06) & 0.76 (+0.07) & 0.62 (+0.01) & 0.32 (+0.05) & 0.74 (+0.06) \\
	LAION + CrossFaceID  & \textbf{0.31 (+0.06)} & \textbf{0.79 (+0.08)} & \textbf{0.63 (+0.01)} & \textbf{0.34 (+0.06)} & \textbf{0.75 (+0.05)} \\\bottomrule
	\end{tabular}

	\caption{Quantitative results of different FaceID customization models on CrossFaceID-test and Unsplash-50, and we highlight the highest score in bold.}
	\label{tab:quantitative_results}
\end{table*}

\begin{table*}[t]
	\centering

	\begin{tabular}{lccccccccc}
	\toprule
	\multirow{2}{*}{\textbf{Model}} & \multicolumn{3}{c}{\textbf{Cusomization}} & \multicolumn{3}{c}{\textbf{Fidelity}} & \multicolumn{3}{c}{\textbf{Quality}} \\
	& \textit{Min} & \textit{Max} & \textit{Avg} & \textit{Min} & \textit{Max} & \textit{Avg} & \textit{Min} & \textit{Max} & \textit{Avg} \\\midrule
	IP-Adapter \cite{ye2023ip} & 0 & 3 & 1.2 & 1 & 3 & 1.7 & 2 & 4 & 3.01 \\
	InstantID \cite{wang2024instantid} & 0 & 3 & 1.64 & 3 & 5 & 4.17 & 3 & 5 & 4.36 \\
	LAION & 0 & 3 & 1.65 & 3 & 5 & \textbf{4.2} & 3 & 5 & 4.38 \\\midrule
	\multicolumn{10}{c}{\textit{Ours (Fine-tuned on CrossFaceID)}} \\
	IP-Adapter + CrossFaceID & 2 & 4 & 2.95 & 1 & 3 & 1.62 & 2 & 4 & 2.98 \\
	InstantID + CrossFaceID  & 3 & 5 & 4.02 & 3 & 5 & 4.17 & 3 & 5 & 4.40 \\
	LAION + CrossFaceID  & \textbf{3} & \textbf{5} & \textbf{4.21} & \textbf{3} & \textbf{5} & \textbf{4.21} & \textbf{3} & \textbf{5} & \textbf{4.43} \\\bottomrule
	\end{tabular}

	\caption{Human evaluations of different FaceID customization models based on three criteria: (1) Customization, (2) Fidelity and (3) Quality, and we highlight the highest score in bold.}
	\label{tab:human_evaluations}
\end{table*}

\subsection{Quantitative Results}

To more effectively evaluate the effectiveness of our CrossFaceID dataset, we conduct quantitative experiments on two test sets: CrossFaceID-test and Unsplash-50 \cite{gal2024lcm}.
Due to the lack of test sets for evaluating the abilities of models in customizing FaceID, we collected CrossFaceID-test.
CrossFaceID-test consists of 200 text-image pairs sourced from the Internet. For each image, we include a version of the same person with a different facial expression or angle, allowing us to assess the models' performance in generating reference images that align with the input text and given face. For Unsplash-50 \cite{gal2024lcm}, it includes 50 text-image pairs, which can be utilized to quantify the models' performance in maintaining FaceID fidelity.

For evaluation metrics, we use the following: (1) Face Sim, which calculates the FaceID cosine similarity between the input face and the face extracted from the generated image, providing a direct estimate of the difference between the generated and input faces. 
It is important to note that Face Sim is evaluated only on the Unsplash-50 test set. This is because, for CrossFaceID-test, the objective is for the model to generate an image with a face that differs from the input face. As a result, Face Sim may not be suitable for assessing this specific goal; 
(2) CLIP-T \cite{radford2021learning}, which evaluates the model's ability to follow prompts; and (3) CLIP-I \cite{radford2021learning}, which measures the CLIP image similarity between the original image and the image after FaceID insertion.

Results are shown in Table \ref{tab:quantitative_results}. From these results, we can observe that after fine-tuning on our CrossFaceID dataset, the model achieves improvement on all metrics. For instance, the CLIP-I metric on the CrossFaceID-test dataset improves from 0.71 (LAION) to 0.79 (LAION + CrossFaceID), while the CLIP-T metric on the Unsplash-50 dataset increases from 0.28 (LAION) to 0.34 (LAION + CrossFaceID). These findings further validate the effectiveness of our CrossFaceID dataset in adhering to user descriptions to both customize the input face and maintain the input face's identity within the generated images.

\subsection{Human Evaluations}

While automated evaluations, as conducted above, effectively measure objective aspects like FaceID fidelity and prompt adherence, they fall short in assessing subjective qualities, such as whether the customized face accurately represents the requested attributes (e.g., expressions or angles) while maintaining resemblance to the input person. 
To address this limitation, we incorporate human evaluations into our experiments. In this way, we collected 200 celebrity faces from the Internet and manually designed prompts to force the evaluated FaceID models to generate images showing different expressions and angles (e.g., smile, sadness, turning head and wearing attire).
The generated images are then evaluated by 10 human participants, who score them based on three criteria: (1) \textbf{Customization:} whether the generated image accurately follows the input prompt to customize the given face, (2) \textbf{Fidelity:} whether the generated image retains the identity of the input face, and (3) \textbf{Quality:} the overall quality of the generated image. 
The scores range from 0 to 5, with 0 indicating the poorest quality and 5 representing the highest quality.

The results are shown in Table \ref{tab:human_evaluations}. From these results, we can observe that: (1) In terms of customization, previous FaceID customization models demonstrate very limited customization capabilities, with average scores of only 1.2 for the IP-Adapter model and 1.64 for the InstantID model. However, after fine-tuning on our CrossFaceID dataset, their customization abilities improve significantly, such as an increase from 1.65 (LAION) to 4.21 (LAION + CrossFaceID). (2) Regarding fidelity, the ability to maintain FaceID fidelity remains stable before and after fine-tuning on our CrossFaceID dataset. For example, 4.17 (InstantID) vs. 4.17 (InstantID + CrossFaceID) and 4.2 (LAION) vs. 4.21 (LAION + CrossFaceID). (3) In terms of quality, the fine-tuning process on our CrossFaceID dataset does not degrade the quality of the generated images but instead slightly improves it, such as an increase from 4.38 (LAION) to 4.43 (LAION + CrossFaceID).

\section{Conclusion}
\label{sec:conclusion}

In this paper, we propose CrossFaceID, the first large-scale, high-quality, and publicly available dataset specifically designed to improve the facial modification capabilities of FaceID customization models. Specifically, CrossFaceID consists of 40,000 text-image pairs from approximately 2,000 persons, with each person represented by around 20 images showcasing diverse facial attributes such as poses, expressions, angles, and adornments. During the training stage, a specific face of a person is used as input, and the FaceID customization model is forced to generate another image of the same person but with altered facial features. This allows the FaceID customization model to acquire the ability to personalize and modify known facial features during the training process, thus improving its FaceID customization abilities during the later inference stage.

We perform comprehensive experiments to demonstrate the effectiveness of our CrossFaceID dataset, revealing that models fine-tuned on this dataset maintain their ability to preserve FaceID fidelity while significantly enhancing their face customization capabilities. Moreover, to support further progress in the FaceID customization domain, we have made our code, datasets, and models publicly available.

{\small
\bibliographystyle{ieee_fullname}
\bibliography{egbib}

\begin{thebibliography}{10}\itemsep=-1pt

\bibitem{achiam2023gpt}
Josh Achiam, Steven Adler, Sandhini Agarwal, Lama Ahmad, Ilge Akkaya, Florencia~Leoni Aleman, Diogo Almeida, Janko Altenschmidt, Sam Altman, Shyamal Anadkat, et~al.
\newblock Gpt-4 technical report.
\newblock {\em arXiv preprint arXiv:2303.08774}, 2023.

\bibitem{balaji2022ediff}
Yogesh Balaji, Seungjun Nah, Xun Huang, Arash Vahdat, Jiaming Song, Qinsheng Zhang, Karsten Kreis, Miika Aittala, Timo Aila, Samuli Laine, et~al.
\newblock ediff-i: Text-to-image diffusion models with an ensemble of expert denoisers.
\newblock {\em arXiv preprint arXiv:2211.01324}, 2022.

\bibitem{chen2023photoverse}
Li Chen, Mengyi Zhao, Yiheng Liu, Mingxu Ding, Yangyang Song, Shizun Wang, Xu Wang, Hao Yang, Jing Liu, Kang Du, et~al.
\newblock Photoverse: Tuning-free image customization with text-to-image diffusion models.
\newblock {\em arXiv preprint arXiv:2309.05793}, 2023.

\bibitem{chen2024dreamidentity}
Zhuowei Chen, Shancheng Fang, Wei Liu, Qian He, Mengqi Huang, and Zhendong Mao.
\newblock Dreamidentity: Enhanced editability for efficient face-identity preserved image generation.
\newblock In {\em Proceedings of the AAAI Conference on Artificial Intelligence}, volume~38, pages 1281--1289, 2024.

\bibitem{dhariwal2021diffusion}
Prafulla Dhariwal and Alexander Nichol.
\newblock Diffusion models beat gans on image synthesis.
\newblock {\em Advances in neural information processing systems}, 34:8780--8794, 2021.

\bibitem{ding2021cogview}
Ming Ding, Zhuoyi Yang, Wenyi Hong, Wendi Zheng, Chang Zhou, Da Yin, Junyang Lin, Xu Zou, Zhou Shao, Hongxia Yang, et~al.
\newblock Cogview: Mastering text-to-image generation via transformers.
\newblock {\em Advances in neural information processing systems}, 34:19822--19835, 2021.

\bibitem{ding2022cogview2}
Ming Ding, Wendi Zheng, Wenyi Hong, and Jie Tang.
\newblock Cogview2: Faster and better text-to-image generation via hierarchical transformers.
\newblock {\em Advances in Neural Information Processing Systems}, 35:16890--16902, 2022.

\bibitem{gal2022image}
Rinon Gal, Yuval Alaluf, Yuval Atzmon, Or Patashnik, Amit~H Bermano, Gal Chechik, and Daniel Cohen-Or.
\newblock An image is worth one word: Personalizing text-to-image generation using textual inversion.
\newblock {\em arXiv preprint arXiv:2208.01618}, 2022.

\bibitem{gal2024lcm}
Rinon Gal, Or Lichter, Elad Richardson, Or Patashnik, Amit~H Bermano, Gal Chechik, and Daniel Cohen-Or.
\newblock Lcm-lookahead for encoder-based text-to-image personalization.
\newblock {\em arXiv preprint arXiv:2404.03620}, 2(3):4, 2024.

\bibitem{gregor2014deep}
Karol Gregor, Ivo Danihelka, Andriy Mnih, Charles Blundell, and Daan Wierstra.
\newblock Deep autoregressive networks.
\newblock pages 1242--1250, 2014.

\bibitem{gulrajani2016pixelvae}
Ishaan Gulrajani, Kundan Kumar, Faruk Ahmed, Adrien~Ali Taiga, Francesco Visin, David Vazquez, and Aaron Courville.
\newblock Pixelvae: A latent variable model for natural images.
\newblock {\em arXiv preprint arXiv:1611.05013}, 2016.

\bibitem{ho2020denoising}
Jonathan Ho, Ajay Jain, and Pieter Abbeel.
\newblock Denoising diffusion probabilistic models.
\newblock {\em Advances in neural information processing systems}, 33:6840--6851, 2020.

\bibitem{huang2023composer}
Lianghua Huang, Di Chen, Yu Liu, Yujun Shen, Deli Zhao, and Jingren Zhou.
\newblock Composer: Creative and controllable image synthesis with composable conditions.
\newblock {\em arXiv preprint arXiv:2302.09778}, 2023.

\bibitem{hurst2024gpt}
Aaron Hurst, Adam Lerer, Adam~P Goucher, Adam Perelman, Aditya Ramesh, Aidan Clark, AJ Ostrow, Akila Welihinda, Alan Hayes, Alec Radford, et~al.
\newblock Gpt-4o system card.
\newblock {\em arXiv preprint arXiv:2410.21276}, 2024.

\bibitem{kim2022diffface}
Kihong Kim, Yunho Kim, Seokju Cho, Junyoung Seo, Jisu Nam, Kychul Lee, Seungryong Kim, and KwangHee Lee.
\newblock Diffface: Diffusion-based face swapping with facial guidance.
\newblock {\em arXiv preprint arXiv:2212.13344}, 2022.

\bibitem{kumari2023multi}
Nupur Kumari, Bingliang Zhang, Richard Zhang, Eli Shechtman, and Jun-Yan Zhu.
\newblock Multi-concept customization of text-to-image diffusion.
\newblock In {\em Proceedings of the IEEE/CVF Conference on Computer Vision and Pattern Recognition}, pages 1931--1941, 2023.

\bibitem{li2024photomaker}
Zhen Li, Mingdeng Cao, Xintao Wang, Zhongang Qi, Ming-Ming Cheng, and Ying Shan.
\newblock Photomaker: Customizing realistic human photos via stacked id embedding.
\newblock In {\em Proceedings of the IEEE/CVF Conference on Computer Vision and Pattern Recognition}, pages 8640--8650, 2024.

\bibitem{nichol2021glide}
Alex Nichol, Prafulla Dhariwal, Aditya Ramesh, Pranav Shyam, Pamela Mishkin, Bob McGrew, Ilya Sutskever, and Mark Chen.
\newblock Glide: Towards photorealistic image generation and editing with text-guided diffusion models.
\newblock {\em arXiv preprint arXiv:2112.10741}, 2021.

\bibitem{peng2024portraitbooth}
Xu Peng, Junwei Zhu, Boyuan Jiang, Ying Tai, Donghao Luo, Jiangning Zhang, Wei Lin, Taisong Jin, Chengjie Wang, and Rongrong Ji.
\newblock Portraitbooth: A versatile portrait model for fast identity-preserved personalization.
\newblock In {\em Proceedings of the IEEE/CVF Conference on Computer Vision and Pattern Recognition}, pages 27080--27090, 2024.

\bibitem{podell2023sdxl}
Dustin Podell, Zion English, Kyle Lacey, Andreas Blattmann, Tim Dockhorn, Jonas M{\"u}ller, Joe Penna, and Robin Rombach.
\newblock Sdxl: Improving latent diffusion models for high-resolution image synthesis.
\newblock {\em arXiv preprint arXiv:2307.01952}, 2023.

\bibitem{radford2021learning}
Alec Radford, Jong~Wook Kim, Chris Hallacy, Aditya Ramesh, Gabriel Goh, Sandhini Agarwal, Girish Sastry, Amanda Askell, Pamela Mishkin, Jack Clark, et~al.
\newblock Learning transferable visual models from natural language supervision.
\newblock In {\em International conference on machine learning}, pages 8748--8763. PMLR, 2021.

\bibitem{raffel2020exploring}
Colin Raffel, Noam Shazeer, Adam Roberts, Katherine Lee, Sharan Narang, Michael Matena, Yanqi Zhou, Wei Li, and Peter~J Liu.
\newblock Exploring the limits of transfer learning with a unified text-to-text transformer.
\newblock {\em Journal of machine learning research}, 21(140):1--67, 2020.

\bibitem{ramesh2022hierarchical}
Aditya Ramesh, Prafulla Dhariwal, Alex Nichol, Casey Chu, and Mark Chen.
\newblock Hierarchical text-conditional image generation with clip latents.
\newblock {\em arXiv preprint arXiv:2204.06125}, 1(2):3, 2022.

\bibitem{ramesh2021zero}
Aditya Ramesh, Mikhail Pavlov, Gabriel Goh, Scott Gray, Chelsea Voss, Alec Radford, Mark Chen, and Ilya Sutskever.
\newblock Zero-shot text-to-image generation.
\newblock In {\em International conference on machine learning}, pages 8821--8831. Pmlr, 2021.

\bibitem{rombach2022high}
Robin Rombach, Andreas Blattmann, Dominik Lorenz, Patrick Esser, and Bj{\"o}rn Ommer.
\newblock High-resolution image synthesis with latent diffusion models.
\newblock In {\em Proceedings of the IEEE/CVF conference on computer vision and pattern recognition}, pages 10684--10695, 2022.

\bibitem{ruiz2023dreambooth}
Nataniel Ruiz, Yuanzhen Li, Varun Jampani, Yael Pritch, Michael Rubinstein, and Kfir Aberman.
\newblock Dreambooth: Fine tuning text-to-image diffusion models for subject-driven generation.
\newblock In {\em Proceedings of the IEEE/CVF conference on computer vision and pattern recognition}, pages 22500--22510, 2023.

\bibitem{saharia2022photorealistic}
Chitwan Saharia, William Chan, Saurabh Saxena, Lala Li, Jay Whang, Emily~L Denton, Kamyar Ghasemipour, Raphael Gontijo~Lopes, Burcu Karagol~Ayan, Tim Salimans, et~al.
\newblock Photorealistic text-to-image diffusion models with deep language understanding.
\newblock {\em Advances in neural information processing systems}, 35:36479--36494, 2022.

\bibitem{song2020denoising}
Jiaming Song, Chenlin Meng, and Stefano Ermon.
\newblock Denoising diffusion implicit models.
\newblock {\em arXiv preprint arXiv:2010.02502}, 2020.

\bibitem{song2020score}
Yang Song, Jascha Sohl-Dickstein, Diederik~P Kingma, Abhishek Kumar, Stefano Ermon, and Ben Poole.
\newblock Score-based generative modeling through stochastic differential equations.
\newblock {\em arXiv preprint arXiv:2011.13456}, 2020.

\bibitem{valevski2023face0}
Dani Valevski, Danny Lumen, Yossi Matias, and Yaniv Leviathan.
\newblock Face0: Instantaneously conditioning a text-to-image model on a face.
\newblock In {\em SIGGRAPH Asia 2023 Conference Papers}, pages 1--10, 2023.

\bibitem{van2016conditional}
Aaron Van~den Oord, Nal Kalchbrenner, Lasse Espeholt, Oriol Vinyals, Alex Graves, et~al.
\newblock Conditional image generation with pixelcnn decoders.
\newblock {\em Advances in neural information processing systems}, 29, 2016.

\bibitem{van2017neural}
Aaron Van Den~Oord, Oriol Vinyals, et~al.
\newblock Neural discrete representation learning.
\newblock {\em Advances in neural information processing systems}, 30, 2017.

\bibitem{vaswani2017attention}
A Vaswani.
\newblock Attention is all you need.
\newblock {\em Advances in Neural Information Processing Systems}, 2017.

\bibitem{wang2024instantid}
Qixun Wang, Xu Bai, Haofan Wang, Zekui Qin, Anthony Chen, Huaxia Li, Xu Tang, and Yao Hu.
\newblock Instantid: Zero-shot identity-preserving generation in seconds.
\newblock {\em arXiv preprint arXiv:2401.07519}, 2024.

\bibitem{xiao2024fastcomposer}
Guangxuan Xiao, Tianwei Yin, William~T Freeman, Fr{\'e}do Durand, and Song Han.
\newblock Fastcomposer: Tuning-free multi-subject image generation with localized attention.
\newblock {\em International Journal of Computer Vision}, pages 1--20, 2024.

\bibitem{xu2024imagereward}
Jiazheng Xu, Xiao Liu, Yuchen Wu, Yuxuan Tong, Qinkai Li, Ming Ding, Jie Tang, and Yuxiao Dong.
\newblock Imagereward: Learning and evaluating human preferences for text-to-image generation.
\newblock {\em Advances in Neural Information Processing Systems}, 36, 2024.

\bibitem{ye2023ip}
Hu Ye, Jun Zhang, Sibo Liu, Xiao Han, and Wei Yang.
\newblock Ip-adapter: Text compatible image prompt adapter for text-to-image diffusion models.
\newblock {\em arXiv preprint arXiv:2308.06721}, 2023.

\bibitem{yuan2023inserting}
Ge Yuan, Xiaodong Cun, Yong Zhang, Maomao Li, Chenyang Qi, Xintao Wang, Ying Shan, and Huicheng Zheng.
\newblock Inserting anybody in diffusion models via celeb basis.
\newblock {\em arXiv preprint arXiv:2306.00926}, 2023.

\bibitem{zhang2023adding}
Lvmin Zhang, Anyi Rao, and Maneesh Agrawala.
\newblock Adding conditional control to text-to-image diffusion models.
\newblock In {\em Proceedings of the IEEE/CVF International Conference on Computer Vision}, pages 3836--3847, 2023.

\bibitem{zhao2023diffswap}
Wenliang Zhao, Yongming Rao, Weikang Shi, Zuyan Liu, Jie Zhou, and Jiwen Lu.
\newblock Diffswap: High-fidelity and controllable face swapping via 3d-aware masked diffusion.
\newblock In {\em Proceedings of the IEEE/CVF Conference on Computer Vision and Pattern Recognition}, pages 8568--8577, 2023.

\end{thebibliography}
}

\end{document}